%% file: main.tex
\documentclass[10pt]{article} 
\usepackage{fancyhdr}
\usepackage[accepted]{tmlr}
\usepackage{mathtools} 
\usepackage{booktabs} 
\usepackage{tikz} 
\usepackage[hyphens]{url}
\usepackage{hyperref}
\usepackage{natbib}
\usepackage{xspace}
\usepackage{graphicx}
\usepackage{pgfplots}
\pgfplotsset{compat=newest}
\usepackage{algpseudocode}
\usepackage{amsmath}
\usepackage{amsfonts}
\usepackage{color}
\usepackage{algorithm}
\newtheorem{definition}{Definition}
 
\newcommand{\ignore}[1]{}
\newcommand{\method}{\textsc{BELLA}\xspace}

\newcommand{\added}[1]{{\textcolor{black}{#1}}}
\newcommand{\newlyadded}[1]{{\textcolor{black}{#1}}}

\def\month{08}  
\def\year{2025} 
\def\openreview{\url{https://openreview.net/forum?id=F9Kv96K}} 

\title{BELLA: Black-box model Explanations\\ by Local Linear Approximations}

\author{\name 	Nedeljko Radulović  \email nedeljko.radulovic88@gmail.com \\
      \addr Telecom Paris,       Institut Polytechnique de Paris, France
      \AND
      \name Albert Bifet \email albert.bifet@telecom-paris.fr \\
      \addr Telecom Paris,       Institut Polytechnique de Paris, France
      \AND
      \name Fabian M. Suchanek \email fabian.suchanek@telecom-paris.fr\\
      \addr Telecom Paris,       Institut Polytechnique de Paris, France
      }
      
\begin{document}

\maketitle

\begin{abstract}
Understanding the decision-making process of black-box models has become not just a legal requirement, but also an additional way to assess their performance. However, the state of the art post-hoc explanation approaches for regression models rely on synthetic data generation, which introduces uncertainty and can hurt the reliability of the explanations. Furthermore, they tend to produce explanations that apply to only very few data points. In this paper, we present BELLA, a deterministic model-agnostic post-hoc approach for explaining the individual predictions of regression black-box models. BELLA provides explanations in the form of a linear model trained in the feature space. BELLA maximizes the size of the neighborhood to which the linear model applies so that the explanations are accurate, simple, general, and robust.
\end{abstract}

\section{Introduction}\label{sec:intro}
Machine Learning (ML) and Artificial Intelligence (AI) models have been employed to handle tasks in various domains, including justice, healthcare, finance, self-driving cars, and many more. 
Consequently, legislative regulations have been proposed to protect interested parties and control the usage of these models. One example is the General Data Protection Regulation of the European Union~\citep{goodman2017european}, which stipulates \textit{the right to an explanation} in situations where an AI system has been employed in a decision-making process. The AI act~\citep{aiact}, too, has stipulated the 
transparency of AI models according to the level of risk they pose. 

The main issue is that many ML models are \textit{black-box models}, i.e. one cannot easily understand how they arrive at a decision.
This has led to the emergence of explainable Artificial Intelligence (xAI), a research field that aims to make black-box models human-understandable.
In this paper, we are concerned with understanding regression models, i.e., models that make a numerical prediction. We are interested in explaining a given prediction of such a model \emph{post-hoc}, i.e., after it has been produced.  
This is usually done by building an interpretable surrogate-model (e.g., a decision tree) that mimics the black-box model and that can be used to understand the prediction.

Numerous approaches have been proposed to build such surrogate models, in particular SHAP~\citep{lundberg2017unified}, LIME~\citep{ribeiro2016should}, and MAPLE~\citep{plumb2018model}. We review them in Section~\ref{related_work}. To evaluate the surrogate models, several criteria have been proposed: we want the surrogate model to be \emph{accurate}, i.e., to reflect the predictions of the black-box model; we want it to be \emph{simple}, i.e., to use few features; we want it to be \emph{robust}, i.e., giving similar explanations to similar data points; and we want it to be \emph{general}, i.e., applicable to many data points. We survey these desiderata in Section~\ref{sec:preliminaries}. We find that existing approaches tend to be good on some of these criteria, but never excel on all of them. This is not surprising, as the desiderata stand in obvious conflict: A simple surrogate model, e.g., risks being not very accurate, because usually accurate predictions can be made only by the type of complex models that we wish to explain in the first place.

Our key idea (which we present in Section~\ref{method}) is to train a local linear model on the neighborhood of the data point that we wish to explain. This allows us to develop an approach called BELLA (Black-box model Explanations by Local Linear Approximations). We can show through extensive experiments (in Section~\ref{experiments}) on a dozen datasets that BELLA beats all existing approaches across nearly all desiderata. 

\section{Related Work}
\label{related_work}

Explainable AI has received much attention in the scientific literature~\citep{beaudouin2020flexible,guidotti2018survey,adadi2018peeking,murdoch2019interpretable,burkart2021survey,hassija2024interpreting}. 
In this paper, we are interested in \emph{post-hoc approaches}, i.e., those that add interpretability to a given black-box model. 
Some of these approaches have been developed specifically for a given type of learners (such as~\cite{fi} for neural models). However, we are interested in \textit{model-agnostic} approaches, i.e., those that can interpret any black-box model. Some model-agnostic approaches compute feature importance~\citep{l2x,vibi}. However, these approaches do not allow explaining unseen data points. Hence, we focus on approaches that build a \emph{surrogate model}, i.e., a model that mimics the black-box model but that is 
interpretable by design (e.g., a decision tree).
While \textit{global} methods provide an interpretation of the black-box model behavior on the whole space, \textit{local} models provide an interpretation for a single data point. In this paper, we are interested \textit{model-agnostic} \emph{post-hoc} \emph{local} explanations for \emph{regression models}, i.e., we aim to provide an explanation for a given real-valued decision by any type of model for a given data point. We are thus not interested in approaches that work for classification only~\citep{aim,dice,copa,l2s}.

One approach to deal with regression models is to adjust the methods for classification models (such as LORE~\citep{guidotti2019factual}), e.g., by discretization or clustering. 
However, this loses information and may require domain knowledge. Therefore several approaches have been developed to natively support both classification and regression models:
SHAP~\citep{lundberg2017unified} introduces a game theory approach to compute the contribution of each feature. The explanation applies to a single data point and it is given as a linear combination of the feature contributions. In order to improve computation time, AcME~\citep{dandolo2023acme} computes feature contributions based on the perturbations based on data quantiles. LIME~\citep{ribeiro2016should} generates synthetic data points by feature perturbations. This yields a weighted neighborhood that is used to train a linear model, whose coefficients are then used as an explanation. However, both LIME and SHAP compute feature contributions in a projected, binary, space, which does not correspond to the original feature space. 
MAPLE~\citep{plumb2018model} addresses this problem and uses Random Forests to assign weights to the training examples. In this way, it forms a weighted neighborhood on which the explanation applies. SHAP, LIME, and MAPLE are direct competitors to our method BELLA, and we will see in our experiments that BELLA outperforms all of them on the quality of the explanations.

DLIME~\citep{zafar2019dlime} is a deterministic variant of LIME that provides stable and consistent explanations. However, it requires extensive manual input, as the user has to provide the number of clusters for the hierarchical clustering step, the number of neighbors for the KNN step of the method, and the length of the explanation. As such, DLIME is not well suited for regression tasks and was thus applied only to classification. 

Another group of approaches computes \textit{counterfactual} explanations. One such method~\citep{white2019measurable} uses the idea of \textit{b-counterfactuals}, i.e., the minimal change in the feature to gauge the prediction of the complex model. This method applies only to classification tasks. Another work~\citep{dandl2020multi} uses Multi-Objective Optimization to compute counterfactual explanations, both for classification and regression. Another work~\citep{mcce} uses Monte Carlo sampling for the same purpose. \added{However, we focus on factual explanations in this paper.}

\section{Preliminaries}\label{sec:preliminaries}

\textbf{Goal.}
We are given a tabular dataset $T \subset F_1 \times ... \times F_n$, where each $F_i$ is a set of \emph{feature values}. \newlyadded{For example, a feature for a real estate dataset could be the size of the lot in square meters, and the set of feature values would then be $\mathbb{R}^+$.} We are also given a function $Y: T \rightarrow \mathbb{R}$ that yields, for each $x \in T$, a \emph{label} $Y(x) \in \mathbb{R}$. \newlyadded{In our example, the labels can be the prices of the houses.} These labels may, e.g., have been produced by a black-box model, in which case the label is a \emph{prediction}. 
Consider now one data point $x \in T$ with its label $Y(x)$. We aim to compute an \emph{explanation} in the following sense~\citep{das2020opportunities}:

\begin{definition}
\label{def:expl}
An explanation is additional meta-information, generated by an external algorithm or by the machine learning model itself, to describe the feature importance or relevance of an input instance towards a particular output classification.
\end{definition} 
\noindent If the label was produced by a black-box model, we cannot be sure post-hoc that the features we identify really contributed to the computation of the label (the model may just as well have thrown a dice, independently of any feature values). However, if several data points with these or similar feature values produce a similar prediction, we can use abductive reasoning to infer that these features may have contributed to the prediction, and that, hence, the features constitute an explanation. This is in fact common in the literature~\citep{ribeiro2016should,lundberg2017unified,radulovic2021confident,ignatiev2019abduction}.

\textbf{Quality measures.}
Several properties of “good” explanations have been proposed. Some of them, such as plausibility and accordance with prior beliefs, require human evaluation. Among the criteria that do not, we commonly find~\citep{miller2019explanation,guidotti2018survey,burkart2021survey,molnar2018guide}:
\begin{enumerate}
    \item \textbf{Fidelity}: we want the value that the surrogate model explains to be close to value that the black-box model predicts.
    \item \textbf{Simplicity}: we want the explanation to contain few features.
    \item \textbf{Robustness}: we want similar data points to have similar explanations.
\end{enumerate}
\noindent In addition, users tend to favor explanations that apply to many data points~\citep{radulovic2021confident}. This appears counter-intuitive, as we aim to explain only a single data point, no matter the others. And yet, it is easy to see that an explanation such as ``You have a high risk of diabetes because your body mass index is 27, your A1C level is 7\%, and your blood sugar level is 210mg/dL'' is little satisfactory, as it allows no generalization. More helpful is to know that, \emph{generally}, people with a body mass index larger than 25, an AIC level above 6.5\%, and a blood sugar level of 200 mg/dL have a high risk of diabetes~\citep{mayo}. We would thus like to have:
\begin{enumerate}
\setcounter{enumi}{3}
 \item \textbf{Generality}: we want the number of data points to which an explanation applies to be large.
\end{enumerate}

\textbf{Additional desiderata.} In addition to the above scalar quality measures, there are also criteria in the literature that either apply or don't apply to a given method of explanation. For example, several methods for post-hoc explainability use randomization to probe the black-box model. However, this entails that the same data can lead to different explanations, which introduces uncertainty for the user~\citep{zhang2019should,slack2020fooling}. 
We thus have as desideratum:
\begin{enumerate}
\setcounter{enumi}{4}
\item \textbf{Determinism}: the avoidance of randomization steps
\end{enumerate}
\noindent Furthermore, some methods~\citep{plumb2018model} propose explanations that take the form of a linear equation, which allows computing the predicted value from the feature values. This is a very attractive property, as the user can toy with the explanation and apply it also to neighboring data points. We thus have a desideratum that we call 
\begin{enumerate}
\setcounter{enumi}{5}
\item \textbf{Verifiability}: the possibility to compute the predicted value from the feature values
\end{enumerate}
\noindent Given this number of quality measures and desiderata, it is not surprising that no existing method (including our own) can satisfy all of them perfectly. However, we can at least show that our method ticks all desiderata, and outperforms existing methods across nearly all quality measures.

\section{BELLA}
\label{method}

We are given a tabular dataset $T$, \newlyadded{real-valued labels $Y: T \rightarrow \mathbb{R}$, and a data point $x \in T$, and we aim to compute an explanation for the label $Y(x)$}. \added{Note that, different from other methods such as LIME and SHAP, we need as input only the dataset $T$ and the labels $Y$ on $T$, and not in addition also the model that generated the labels $Y$. This means that, different from SHAP and LIME, our method can explain any labeled dataset, whether the labels were produced by a model or whether they come from any other source (e.g., housing prices from real estate data).} \newlyadded{If BELLA is to explain a model, we first run the model on the training dataset. We then use as labels $Y$ not the ground truth from the training dataset, but the predictions of the model.}

To explain the label $Y(x)$, our idea is to find a linear equation  $Y(x) \approx w_1 \cdot x_1 + w_2 \cdot x_2 + \dots + w_l \cdot x_l + w_0$, where $w_i$ are real-valued regression coefficients and $x_i$ are feature values of $x$ in $T$. Such an equation tells the user (1) what the important features are and (2) how they can be used to compute the predicted value. To find this equation, 
BELLA proceeds in three steps (Algorithm~\ref{alg1}):
\begin{itemize}
    \item[1.] Compute the distance of $x$ to the other points in $T$.
    \item[2.] Conduct a linear search to find the best neighborhood of $x$, according to a defined metric.
    \item[3.] Train a sparse linear model on that neighborhood, and propose this model
    as an explanation.
\end{itemize}
\begin{algorithm}[H]
  \begin{algorithmic}[1]
  \Statex \textbf{Input}: Dataset $T$ with labels $Y$
  \Statex \hspace*{11.3mm} Labeled data point $x \in T$
  \State $d \gets \textrm{ComputeDistances}(x, T)$
  \State $L, N \gets \textrm{NeighborhoodSearch}(x, T, d)$
     \State \Return $L, N$ 
    \end{algorithmic}
      \caption{BELLA}    
      \label{alg1}
\end{algorithm}
\textbf{Step 1: Computing the distances.} To compute the neighborhood of the input data point, we need a distance measure. A good starting point is to have all numerical features on the same scale so that each feature contributes to the distance measure in the same range. Therefore, we first standardize all numerical features to have a mean of 0 and a standard-deviation of 1. 

To compute the distances, we employ the generalized distance function~\citep{harikumar2015k}, which consists of three separate distance measures to account for numerical, categorical, and binary data types, as follows:
\begin{align}
    d(x, x') = & \sum_{i=1}^{m_{\textit{num}}} d_{\textit{num}}(x_{i}, x'_{i} ) + \sum_{j=m_{\textit{num}} + 1}^{m_c + m_{\textit{num}}} d_{c}(x_{j}, x'_{j} ) + \nonumber\\
    &\sum_{k=m_c + m_{\textit{num}} + 1}^{m_{\textit{num}} + m_c + m_b} d_{b} (x'_{k}, x'_{k})
\label{eq:distance}
\end{align}
\noindent Here, $m_{\textit{num}}$, $m_c$ and $m_b$ are the number of numerical, categorical, and binary features, respectively.  
The distance measure for the numerical attributes $d_{\textit{num}}$ is the $L1$ norm $d_{\textit{num}}(x, x' ) = |x - x'|$, which is preferred over $L2$, as it is more robust to outliers~\citep{hopcroft2014foundations}.
\noindent For categorical features, $d_{c}$ is the distance measure~\citep{ahmad2007k}, which takes into account the distribution of values and their co-occurrence with values of other attributes. The distance between two values $a$ and $a'$ of an attribute $A_i$ with respect to attribute $A_j$ is given by:
\begin{equation*}
    d_c^{ij}(a, a') = P(A_j \in \omega | A_i=a) + P(A_j \not\in \omega | A_i=a') - 1
\end{equation*}
\noindent Here, $P(A_j \in \omega | A_i=a)$ is the conditional probability that attribute $A_j$ will take a value from the set $\omega$ given that the attribute $A_i$ takes the value $a$. $\omega$ is a subset of all possible values of attribute $A_j$ that maximizes the sum of the probabilities. Since both probabilities can take values from $[0, 1]$, we subtract 1 in order to arrive at $d_c^{ij}(x,x') \in [0,1]$. Lastly, for binary features, we use the Hamming distance: $d_{h}(x, x')=1$ if $x = x'$, and zero otherwise. In Line $1$ of Algorithm~\ref{alg1}, the function \textit{ComputeDistances} returns the distances by Equation~\ref{eq:distance}. \added{Note that our distance measure does not take the label into account. This is because there can be data points with different feature values and a similar prediction. In such cases, BELLA provides the explanation for the “correct” local neighborhood.}

\textbf{Step 2: Neighborhood Search.} After computing the distances, we proceed with the exploration of the neighborhood of the input data point $x$. The goal is to find a set of points, closest to $x$ according to the distance measure, that will serve as a training set for a local surrogate model. Several common techniques could be considered to that end, including kNN, K-Means, and other distance-based clustering methods. In our case, however, we aim to find a neighborhood such that a linear regression model trained on that neighborhood represents an accurate local approximation of the black-box model. Hence, the quality of the neighborhood is proportional to the quality of the performance of the linear model fitted on it. Common drawbacks of regression evaluation metrics are missing interpretability, sensitivity to outliers and near-zero values, divisions by zero, missing bounds, and missing symmetry. We find that the \textit{Berry-Mielke universal R value} $\Re$~\citep{berry1988generalization} avoids most of these pitfalls. $\Re$ represents the measure of agreement between raters and it is a generalization of Cohen's kappa~\citep{cohen1960coefficient}. 
$\Re$ measures how much better the model is compared to a naive one (e.g., to a random predictor). $\Re$ takes values from the range $[0,1]$, and it can be interpreted easily: If $\Re$ is equal to $0$, the model performance is equal to the one of the random model and if it is $1$, then the model has perfect performance. $\Re$ is defined as $\Re = 1 - \frac{\delta}{\mu}$, 
where $\delta$ and $\mu$ are defined as:
\begin{equation}
     \delta = \frac{1}{N} \sum_{i=1}^{N} \Delta(\hat{y}_{i}, y_i),~~
    \mu = \frac{1}{N^2} \sum_{i=1}^{N} \sum_{j=1}^{N} \Delta(\hat{y}_j, y_i).
    \label{eq:delta_mu}
\end{equation}
Here, $N$ is the number of samples, $y_i$ is the actual label, $\hat{y}_i$ is the predicted value, and $\Delta(\cdot)$ represents the distance function between the true and the predicted value. The original work by~\citep{berry1988generalization} uses the Euclidean distance, but later works~\citep{janson2001measure,janson2004measure} propose to use the squared Euclidean distance instead, because this distance is equivalent to the variance of the variable, which further improves the interpretability of $\Re$. We follow this argumentation, and use $\Delta(a, b) = (a - b)^2$. This definition implies that $\Delta$ is in fact equal to the Mean Squared Error (MSE). Thus, by optimizing $\Re$, we are actually optimizing the accuracy of the local model. 

\begin{figure}
\centering
    \scalebox{0.45}{\input{bad_explanation.tikz}}
    \scalebox{0.45}{\input{good_explanation.tikz}}
    \caption{Left: an explanation for a data point $x$ that is too specific, applying only to a very small neighborhood. Right: An explanation that applies to a larger neighborhood, which is what we aim at.}
    \label{fig:generality}
\end{figure}

However, to avoid explanations that are too specific, i.e., explanations that apply to very small neighborhoods, as in Figure~\ref{fig:generality} (left), we wish to optimize not just the \emph{accuracy}, but also the \emph{generality} of the surrogate model. Therefore, we include the size of the neighborhood in the optimization function, to aim for explanations that are at the same time accurate and general (Figure~\ref{fig:generality} (right)). One way to do this is to maximize the lower bound of the confidence interval of $\Re$. 
The lower and upper bounds for the confidence interval of $\Re$ are given by~\citep{berry1988generalization}: 
\begin{equation}
    CI_{\Re} = \overline{\Re} \pm \textit{MOE}_{\Re} = 1- \frac{\overline{\delta} \mp \textit{MOE}_{\delta}}{\mu}
    \label{eq:moe}
\end{equation}
\noindent Here, \textit{MOE} stands for the Margin of Error. From Equation~\ref{eq:moe}, it follows that computing the lower bound of $\Re$ is analogous to computing the upper bound of $\delta$. Therefore, we can compute the margin of error for $\delta$ as $\textit{MOE}_{\delta} = t\frac{\sigma}{\sqrt{N}}$, where $\sigma$ is the standard deviation of the sample, $N$ is the sample size and $t$ represents the critical value from the t-distribution. We use the t-distribution because it is adapted for small sample sizes, which is what we encounter when we grow the neighborhood. 
The distribution converges to the normal distribution as the sample size increases. 

Due to the non-monotonic nature of the $\Re$ value, we have to explore the whole space to maximize its lower bound. We employ a linear search algorithm (Algorithm~\ref{alg3}) to this end. 
\begin{algorithm}
  \begin{algorithmic}[1]
  \Statex \textbf{Input}: Labeled data point $x \in T$
  \Statex \hspace*{11.3mm} Dataset $T$ with labels $Y$ 
  \Statex \hspace*{11.3mm} Distances $d\!:\!T\!\!\rightarrow\mathbb{R}$ of the data points to $x$
  \State Sort $T$ by ascending $d$
  \State $n \gets$ number of features in $T$
  \State $max\Re_{lb} \gets 0, \textit{bestN} \gets 0, \textit{bestL} \gets \emptyset$ 
  \For{$i=\textit{min}(2n,|T|)$ to $|T|$} 
    \State $L \gets \textit{TrainLocalSurrogateModel}(T[0:i])$
    \If {$\Re_{lb}(L) > max\Re_{lb}$}
    \State {
    $max\Re_{lb} \gets \Re_{lb}(L)$, $\textit{bestN}\!\gets\!{}i$, $\textit{bestL}\!\gets\!{}L$
    }
    \EndIf
  \EndFor
     \State \Return \textit{bestL, T[0:bestN]}
     \end{algorithmic}
       \caption{Neighborhood Search}
  \label{alg3}
\end{algorithm}

\noindent The algorithm receives as input a labeled data point $x$ that is to be explained, a labeled training set $T$, and a vector of distances between $x$ and each point in the training set $T$. We sort the training set by increasing distance to $x$, train a linear model on the first $i$ data points for increasing $i$, 
and return the set of neighbors for which the lower bound of $\Re$ is maximal. As the neighborhood is very small in the beginning, the training easily leads to overfitting.
Therefore, we consider at least 2$n$ data points for our neighborhood, where $n$ is the number of features. This ensures that the estimation of regression coefficients exhibits less than $10\%$ relative bias~\citep{austin2015number}.

\begin{algorithm}
  \begin{algorithmic}[1]
  \Statex \textbf{Input}: Neighborhood of data points $\mathcal{N}$ 
  \State $F \gets $ the set of all features in $\mathcal{N}$
  \State $F' \gets \{f | f \in F \wedge \textit{VIF}(f) < 10.0\}$ 
  \State $\textit{Features}_{Lasso}  \gets Lasso(cv=5, \textit{features}=F')$
   \State \Return $OLS(\textit{Features}_{Lasso})$
    \end{algorithmic}
      \caption{Train Local Surrogate Model}
  \label{alg2}
\end{algorithm}

\textbf{Step 3: Building a local surrogate model.} 
We build our local surrogate model on the neighborhood we have found.
To obtain a model with few parameters (i.e., a \emph{simple} model), we use regularization. In terms of feature selection, $L1$ regularization (e.g. Lasso~\citep{hastie2009elements}) is able to select a nearly perfect subset of variables in a wide range of situations. The only condition for this to work is that there are no highly collinear variables~\citep{candes2009near}, which can significantly reduce the precision of estimated regression coefficients. To remove highly collinear features, we compute the \textit{variance inflation factor} (VIF), and, following a rule of thumb~\citep{stine1995graphical}, adopt $10$ as the cut-off value for the VIF.

After removing highly collinear features, the next step is to train a linear model with Lasso regularization. 
Lasso regularization adds a penalty term in the form of the sum of absolute values of the regression coefficients. The objective function is $\min_{\beta \in \mathbb{R}^p}(||y - \beta X||_{2}^{2} + \lambda ||\beta||_1)$, where $\lambda$ is the shrinkage parameter. This provides a sparse model, by forcing some coefficients to be zero. Removing some features ensures a better generalization, and results in simpler, and thus more comprehensible explanations. On the other hand, coefficients obtained by minimizing the Lasso objective function are biased towards zero. Therefore, Lasso is preferred for model selection rather than for prediction. The common strategy is to train an Ordinary Least Squares (OLS) linear model on the subset of variables selected by Lasso. This corresponds to a special variation of the relaxed Lasso~\citep{meinshausen2007relaxed}, with $\phi = 0$.

To determine the value of the shrinkage coefficient $\lambda$, we use 5-fold cross-validation (CV). To preserve the deterministic nature, we perform CV on adjacent slices of the dataset, without random shuffles. CV selects the best model in terms of the prediction error. Since the goal of this step is model selection, we want to avoid choosing $\lambda$ too small, and hence we apply the common one-standard error rule. According to this rule, the most parsimonious model is the one whose error is no more than one standard error above the error of the best model~\citep{hastie2009elements}.

Once we have obtained the most parsimonious model, i.e., the best set of features, we train the final local surrogate model as an OLS model using the features selected by Lasso. This procedure is described in Algorithm~\ref{alg2}, and returns a local linear model. 
\added{The method is designed to maximize the robustness of our explanations: the Lasso regularization makes our results less brittle. Also, our target metric increases the number of data points, making the result more general and thus more robust.}

\textbf{Providing an explanation.} As the final result, BELLA outputs the OLS model computed by Algorithm~\ref{alg2}, together with the size of the neighborhood. As an example, consider the Iranian Churn dataset~\citep{jafari2020optimum}. It contains the (anonymized) customers of a telecommunication company, with their age, subscription length, satisfaction with the service, etc. The goal is to predict the commercial value of the customer to the company (in dollars). Let us now consider a given customer, for whom a black-box model predicted a commercial value of \$551. The explanation that BELLA can provide for this prediction is shown in Figure~\ref{fig:exp_example}. 
\begin{figure}
    \centering
        \caption{Explanation example.}
     \input{exp_example2.tikz}
    \label{fig:exp_example}
\end{figure}
All numerical features have been standardized to have a mean value equal to $0$ and a standard deviation equal to $1$. (Thus, a customer has a ``negative age'' if they are younger than the average customer.) In the explanation, the base value is the output of the model when all inputs are set to zero (i.e. to their mean value). Each bar shows the total contribution of each feature to the predicted value -- \added{positive contributions are the blue upward arrows, and negative contributions are the red downward arrows}. The more the customer phones (variable \emph{seconds}), the more revenue the company generates. The age (which is below average for this particular customer), likewise, has a small positive impact. 
The number of SMS, in contrast, (variable \emph{freqSMS}) impacts the revenue negatively. Finally, the number of distinct phone numbers called~(variable \emph{distnum}) has a small negative impact.
These sizes of the bars are easy to interpret: The size of each bar is equal to the value of the feature multiplied by the weight computed by our method. Their sum is then directly equivalent to the explained value:
\begin{equation*}
    \begin{split}
     y \approx 458.47 + 190.27 \times \textit{seconds} - 102.91 \times \textit{age} - 480.08 \times \textit{freqSMS} - 17.71 \times \textit{distNums}
    \end{split}
    \label{eq:lin_model}
\end{equation*}
This computation applies to all data points in the neighborhood of the input data point (to the current instance and 476 others in our example). We thus see that BELLA's explanations are \emph{verifiable} (because they take the form of a linear equation), \emph{deterministic} (because BELLA does not use any randomized steps), \emph{simple} (because we applied regularization), \emph{general} (because we maximized the neighborhood), and \emph{accurate} (because we optimized the linear model on the local neighborhood). In addition, BELLA does not probe the black-box model. This means that, unlike many of its competitors, BELLA can explain not just the decisions of a black-box model, but any numerical variable in a tabular dataset -- even if that variable was not generated by a model at all but merely observed in reality.

\begin{table}[b]
    \caption{Regression Datasets}
    \label{tab:datasets}
\centering
\begin{tabular}{lrrrrr}
\toprule
 \multicolumn{2}{l}{Dataset~~~~~~~~~~~~~Features} & Numerical & Categorical & Instances\\
 \midrule
 Auto MPG  & $7$ & $6$ & $1$ & $392$\\ 
 Bike  & $12$ & $9$ & $3$  & $8760$\\
 Concrete  & $8$ & $ 8 $ & $ 0 $ & $1030$\\
 Servo & $4$ & $0$ & $4$ & $167$\\
 Electrical & $12$ & $12$ & $0$ & $10000$\\
 Superconductivity & $81$ & $81$ & $0$& $21262$\\
 White Wine Quality& $11$ & $11$& $0$ & $4898$\\ 
 Real Estate Valuation& $5$ & $5$ & $0$ & $414$\\
 Wind & $14$ & $14$ & $0$ & $6574$\\
 CPU activity & $12$ & $12$ & $0$ & $8192$\\
 Echocardiogram & $9$ & $6$ & $3$ & $17496$\\
 Iranian Churn & $11$ & $8$ & $3$ & $3150$\\
\bottomrule
\end{tabular}
\end{table}

\section{Experiments}
\label{experiments}
\textbf{Datasets.} We performed experiments on datasets from two standard repositories~\citep{uci,romano2021pmlb} (shown in Table~\ref{tab:datasets}). Among them is also a high-dimensional dataset, Superconductivity, with 81 features. All categorical features have been one-hot encoded and all numerical features have been standardized. \added{We draw a random 10\% of each dataset as testing data}.
To show that BELLA works with different families of models, we trained a random forest (\added{with 100 trees}), and a neural network (with one hidden layer with 500 nodes) as black-box models. Since the results do not differ much, we show only experiments with the neural network here, while the experiments with the random forest are in Appendix~\ref{sec:trees}.

\textbf{BELLA.} 
Our method is implemented in Python. \added{We set the step size to 10\%.} For the black-box models, we use the implementations of \texttt{scikit-learn}~\citep{scikit-learn}. All experiments are run on a Fedora Linux (release 38) computer with an Intel(R) Xeon(R) v4 @ 2.20GHz CPU, a memory of 64 GB, and Python 3.9.  
All code and the data for BELLA and the experiments is available on Github (URL masked for anonymity).

\textbf{Competitors.} We compare BELLA to LIME~\citep{ribeiro2016should}, SHAP~\citep{lundberg2017unified} and MAPLE~\citep{plumb2018model}. We use the implementations by the authors\footnote{https://github.com/marcotcr/lime}\footnote{https://github.com/slundberg/shap}\footnote{https://github.com/GDPlumb/MAPLE/}. We do not compare to methods that are designed for classification tasks, or that can provide only counterfactual explanations and not factual ones (see again Section~\ref{related_work}). 

\subsection{Experimental results}

\newcommand{\conf}[1]{\textcolor{gray}{\fontsize{6}{6}\selectfont$\pm$#1}}

\begin{table}
    \caption{Fidelity comparison~(RMSE -- smaller is better)}     \label{tab:fidelity_rmse}
\centering
\begin{tabular}{lr@{}lr@{}lr@{}lr}
 \toprule
 Dataset & \multicolumn{2}{c}{LIME} & \multicolumn{2}{c}{MAPLE} & \multicolumn{2}{c}{BELLA} & SHAP\\
  \midrule
 Auto MPG & 2.99 &\conf{0.830} & \textbf{0.86}&\conf{0.270} & 1.45&\conf{0.390} & \textbf{0.00} \\ 
 Bike & 579.76&\conf{24.73} & \textbf{75.42}&\conf{6.710} & 224.70&\conf{12.90} & \textbf{0.00} \\
 Concrete & 10.61&\conf{1.410} & \textbf{2.13}&\conf{0.290} & 4.87&\conf{0.670} & \textbf{0.00} \\
 Servo  & 0.75&\conf{0.320} & \textbf{0.21}&\conf{0.100} & 0.59&\conf{0.260} & \textbf{0.00} \\
 Electrical & 0.02&\conf{0.002} & \textbf{0.01}&\conf{0.001} & 0.02&\conf{0.002} & \textbf{0.00} \\
 Supercond. & 23.17&\conf{0.697} & \textbf{1.05}&\conf{0.029} & 14.24&\conf{0.434} & \textbf{0.00} \\
 White Wine& 0.36&\conf{0.030} & \textbf{0.17}&\conf{0.010} & 0.29&\conf{0.020} & \textbf{0.00} \\ 
 Real Estate& 4.97&\conf{1.170} & \textbf{1.48}&\conf{0.540}& 2.01&\conf{0.730} & \textbf{0.00} \\
 Wind & 2.52&\conf{0.382} & \textbf{1.15}&\conf{0.173} & 1.69&\conf{0.247} & \textbf{0.00} \\
 CPU Activity & 16.30&\conf{1.620} & \textbf{0.81}&\conf{0.060} & 1.18&\conf{0.110} & \textbf{0.00}\\
 Echocard. & 3.02&\conf{0.046} & \textbf{1.82}&\conf{0.031} & 2.84&\conf{0.049} & \textbf{0.00} \\
 Iranian Churn  & 172.13&\conf{23.52} & \textbf{4.04}&\conf{0.970} & 24.40&\conf{7.950} & \textbf{0.00}\\
 \midrule
\textbf{Norm. avg.} & 0.10&\conf{0.014} & 0.02&\conf{0.004} & 0.05&\conf{0.008} & 0.00\\
\bottomrule
\end{tabular}
\end{table}

We compare BELLA's performance against the competitors on the quality measures from Section~\ref{sec:preliminaries}. All tables show the average performance on the test set of each method with confidence intervals at $\alpha=95\%$.

\textbf{Fidelity} is measured by the Root Mean Squared Error (RMSE) of the local surrogate models wrt. the predictions of the black-box models (Table~\ref{tab:fidelity_rmse}, with a min-max normalized average). SHAP always has an error of 0. This is because it provides exact explanations that apply only to a single data point. Among the methods that apply to a neighborhood of points, MAPLE is constantly the best, followed closely by BELLA.\footnote{\added{The fidelity of BELLA could be improved by giving more weight to the explained examples, as MAPLE does, but this would compromise the advantage of BELLA of providing a linear explanation that is valid for the whole neighborhood with the same error margin.}} LIME comes last.

\begin{table}
    \caption{Generality comparison~(\% - larger is better)}\label{tab:generality}
\centering
    \begin{tabular}{lr@{}lrr@{}lr@{}l}
 \toprule
 Dataset & \multicolumn{2}{c}{LIME} & SHAP & \multicolumn{2}{c}{MAPLE} & \multicolumn{2}{c}{BELLA}\\
  \midrule
 Auto MPG & 21.12&\conf{6.120} & 0.00 & \textbf{60.09}&\conf{9.030} & 44.21&\conf{3.210}\\ 
 Bike & 1.34&\conf{0.140} & 0.00 & 6.35&\conf{0.075} & \textbf{45.08}&\conf{2.020}\\
 Concrete & 1.10&\conf{0.420} & 0.00 & 32.05&\conf{1.100} & \textbf{34.02}&\conf{4.140}\\
 Servo & 13.43&\conf{5.040} & 0.00 & \textbf{76.23}&\conf{6.330} & 75.28&\conf{13.21}\\
 Electrical  & 0.02&\conf{0.010} & 0.00 & 8.44&\conf{0.390} & \textbf{31.84}&\conf{0.654}\\
 Supercond. & 1.01&\conf{0.071} & 0.00 & 4.39&\conf{0.070} & \textbf{52.78}&\conf{0.124}\\
 White Wine& 0.97&\conf{0.024} & 0.00 & 16.45&\conf{0.340} & \textbf{33.68}&\conf{3.075}\\ 
 Real Estate & 3.43&\conf{2.120} & 0.00 & \textbf{47.89}&\conf{3.790} & 39.37&\conf{11.12}\\
 Wind  & 0.78&\conf{0.190} & 0.00 & 12.46&\conf{0.540} & \textbf{100.0}&\conf{0.000}\\
 CPU Activity & 0.79&\conf{0.130} & 0.00 & 9.34&\conf{0.215} & \textbf{30.36}&\conf{1.780}\\
 Echocard. & 0.06&\conf{0.003} & 0.00 & 9.19&\conf{0.060} & \textbf{77.11}&\conf{7.270}\\
 Iranian Churn & 1.83&\conf{0.210} & 0.00 & 12.17&\conf{0.410} & \textbf{30.21}&\conf{2.890}\\
 \midrule
 \textbf{Average} & 3.82&\conf{1.206} & 0.00 & 24.59&\conf{1.863} & \textbf{49.50}&\conf{4.119}\\
\bottomrule
\end{tabular}%
\end{table}

\textbf{Generality} is measured by the number of data points to which the explanation applies (as a percentage of all data points in the training set). \added{One could give a hypercube for the size of the neighborhood, but it is arguably the number of data points (and not the size of a potentially sparse hypercube) that conveys the significance of the explanation.}
Thus, for BELLA, we simply return the size of the neighborhood. For MAPLE we return the number of data points that have weights larger than $0$. For LIME, an explanation comes with the range of values for each feature. We count the number of data points that fall into this range. The results are shown in Table~\ref{tab:generality}. For SHAP, the size of the neighborhood is always 0. This is because SHAP provides feature contributions that are specific for the given data point, and there is no way to apply these explanations to other data points. LIME's explanations are more general, and MAPLE's explanations even more. Still, they are vastly less general than the explanations of BELLA. 

\textbf{Simplicity} is most commonly measured by the number of features that an explanation contains (Table~\ref{tab:simplicity}). LIME has the same size of explanations as BELLA. This is because LIME takes this parameter as input and we set it to the size of the explanation provided by BELLA. SHAP and MAPLE constantly provide longer explanations than  BELLA. MAPLE has higher complexity than SHAP, even though it comes with lower accuracy. 

\added{One could consider tuning the simplicity of LIME until LIME beats BELLA on fidelity, or tune fidelity until LIME beats BELLA on simplicity. To compare the two methods, however, one has to fix one parameter and compare the other. This is what our experiments do: at the same simplicity, BELLA beats LIME on fidelity (Table~\ref{tab:fidelity_rmse}). It follows that, to achieve the same fidelity as BELLA, LIME necessarily has to decrease its simplicity. Thus, BELLA beats LIME in both cases.}

\begin{table}
    \caption{Simplicity comparison (smaller values are better). LIME requires the explanation size as input, and we give it the size of the explanation computed by BELLA. }     \label{tab:simplicity}
\centering
    \begin{tabular}{lr@{}lr@{}lr@{}l}
 \toprule
 Dataset & \multicolumn{2}{c}{SHAP} & \multicolumn{2}{c}{MAPLE} & \multicolumn{2}{l}{BELLA/LIME}\\
  \midrule
{Auto MPG} & 9.00&\conf{0.000} & 8.93&\conf{0.090} & \textbf{3.90}&\conf{0.310}\\ 
{Bike} & 11.54&\conf{0.040} & 13.44&\conf{0.080} & \textbf{8.47}&\conf{0.110} \\
{Concrete} & 8.00&\conf{0.000} & 8.00&\conf{0.000} & \textbf{6.24}&\conf{0.240}\\
{Servo}  & 12.47&\conf{0.320} & 17.88&\conf{0.170} & \textbf{5.65}&\conf{1.160}\\
{Electrical} & 12.00&\conf{0.000} & 12.00&\conf{0.000} & \textbf{9.40}&\conf{0.184} \\
{Supercond.}  & 70.29&\conf{0.221} & 81.00&\conf{0.000} & \textbf{14.19}&\conf{0.182}\\
{White Wine} & 11.00&\conf{0.000} & 11.00&\conf{0.000} & \textbf{7.58}&\conf{0.200}\\
{Real Estate~~~~~}  & 5.00&\conf{0.000} & 5.00&\conf{0.000} & \textbf{4.10}&\conf{0.320}\\
{Wind} & 13.05&\conf{0.213} & 14.00&\conf{0.000} & \textbf{9.32}&\conf{0.093}\\
{CPU Activity} & 12.00&\conf{0.000} & 12.00&\conf{0.000} & \textbf{9.56}&\conf{0.190}\\
{Echocardiogram} & 7.33&\conf{0.110} & 8.49&\conf{0.121} & \textbf{7.07}&\conf{0.090}\\
{Iranian Churn} & 9.14&\conf{0.040} & 10.52&\conf{0.060} & \textbf{4.76}&\conf{0.160}\\
   \midrule
   {\textbf{Norm. Avg.}} & 0.89&\conf{0.004} & 0.96&\conf{0.003} & \textbf{0.59}&\conf{0.023}\\
\bottomrule
\end{tabular}
\end{table}

\textbf{Robustness} judges how similar the explanations for close data points are. We measure robustness as:
\begin{equation}
    \textit{robustness} = 1-\frac{1}{n}\sum_{i=1}^n \frac{|w_{1i} - w_{2i}|}{|w_{1i}| + |w_{2i}|}.
\end{equation}
\noindent Here, $n$ is the number of features, and $w_{1i}$ and $w_{2i}$ are the weights of feature $i$ in the first and second explanation, respectively.
Robustness is in the range of $[0,1]$, with 1 indicating that two explanations are identical.
We compute explanations for each data point in the test set, and compute robustness wrt. \added{a smaller set of 5 closest neighbors, and a larger set of 20 closest neighbors (Table~\ref{tab:robustness}). As expected, all methods are a bit less robust when the set of neighbors is larger, but otherwise the results are very similar: }
LIME samples $5000$ data points to create a synthetic neighborhood. Thus, LIME can perform better than our approach \added{on datasets that have fewer observations. However, in the vast majority of cases, as well as on average, BELLA outperforms LIME.}  BELLA also outperforms SHAP by a wide margin. This is because SHAP's explanations are tailored for a single data point. BELLA also outperforms MAPLE. This is because the crisp neighborhood of BELLA provides much more robust explanations than MAPLE's weighted neighborhood. 

\begin{table}
    \caption{\added{Robustness comparison~(0 to 1 -- larger is better)}} \label{tab:robustness}
\centering
\scalebox{0.8}{%
    \begin{tabular}{lr@{}lr@{}lr@{}lr@{}lr@{}lr@{}lr@{}lr@{}lr@{}lr@{}lr@{}lr@{}l}
 \toprule
 Dataset  & \multicolumn{8}{c}{\textbf{Number of Neighbors = 5}}& \multicolumn{8}{c}{\textbf{Number of Neighbors = 20}} \\
 \cmidrule(lr){2-9} \cmidrule(lr){10-17}
          & \multicolumn{2}{c}{LIME} & \multicolumn{2}{c}{SHAP} & \multicolumn{2}{c}{MAPLE} & \multicolumn{2}{c}{BELLA} & \multicolumn{2}{c}{LIME} & \multicolumn{2}{c}{SHAP} & \multicolumn{2}{c}{MAPLE} & \multicolumn{2}{c}{BELLA} \\
  \midrule
{Auto MPG} & 0.89&\conf{0.050} & 0.74&\conf{0.070} & 0.73&\conf{0.060} & \textbf{0.91}&\conf{0.040} & 0.82&\conf{0.053} & 0.69&\conf{0.065} & 0.66&\conf{0.034} & \textbf{0.85}&\conf{0.077}\\ 
{Bike} & 0.78&\conf{0.030} & 0.67&\conf{0.040} & 0.52&\conf{0.040}& \textbf{0.83}&\conf{0.050} & \textbf{0.82}&\conf{0.027} & 0.63&\conf{0.032} & 0.52&\conf{0.047}& 0.81&\conf{0.035}\\
{Concrete} & 0.79&\conf{0.050} & \textbf{0.81}&\conf{0.030} & 0.68&\conf{0.060} & 0.74&\conf{0.070}& 0.63&\conf{0.050} & \textbf{0.70}&\conf{0.047} & 0.64&\conf{0.044} & 0.64&\conf{0.058}\\
{Servo} & \textbf{0.85}&\conf{0.040} & 0.46&\conf{0.020} & 0.53&\conf{0.110} & 0.76&\conf{0.070} & \textbf{0.88}&\conf{0.339} & 0.42&\conf{0.313} & 0.55&\conf{0.161} & 0.78&\conf{0.041}\\
{Electrical} & 0.65&\conf{0.029} & 0.58 &\conf{0.041}& 0.69&\conf{0.045} & \textbf{0.83}&\conf{0.031} & 0.62&\conf{0.015} & 0.55 &\conf{0.022}&0.74&\conf{0.034} & \textbf{0.83}&\conf{0.022}\\
{Supercond.} & \textbf{0.91}&\conf{0.017} & 0.80&\conf{0.036} & 0.60&\conf{0.076} & \textbf{0.91}&\conf{0.036} & 0.90&\conf{0.030} & 0.74&\conf{0.044} & 0.49&\conf{0.061} & \textbf{0.92}&\conf{0.044}\\
{White Wine} & 0.72&\conf{0.060} & 0.64&\conf{0.060} & 0.71&\conf{0.060} & \textbf{0.77}&\conf{0.070} & 0.59&\conf{0.041} & 0.54&\conf{0.042} & 0.64&\conf{0.040} & \textbf{0.67}&\conf{0.047}\\
{Real Estate~~~~~} & 0.77&\conf{0.060} & 0.75&\conf{0.060} & \textbf{0.78}&\conf{0.060} & \textbf{0.78}&\conf{0.090} & 0.67&\conf{0.046} & 0.65&\conf{0.051} & 0.62&\conf{0.059} & \textbf{0.77}&\conf{0.044}\\
{Wind} & 0.68&\conf{0.059} & 0.59&\conf{0.039} & 0.66&\conf{0.028} & \textbf{0.99}&\conf{0.007} & 0.72&\conf{0.085} & 0.56&\conf{0.047} & 0.62&\conf{0.025} & \textbf{0.98}&\conf{0.016}\\
{CPU Activity} & 0.53&\conf{0.080} & 0.73&\conf{0.060} & 0.70&\conf{0.040} & \textbf{0.82}&\conf{0.050} & 0.42&\conf{0.038} & 0.71&\conf{0.050} & 0.75&\conf{0.025} & \textbf{0.81}&\conf{0.035}\\
{Echocardiogram} & 0.77&\conf{0.031} & 0.64&\conf{0.038} & 0.61&\conf{0.041} & \textbf{0.92}&\conf{0.057} & 0.76&\conf{0.037} & 0.63&\conf{0.036} & 0.57&\conf{0.033} & \textbf{0.84}&\conf{0.070}\\
Iranian Churn & 0.71&\conf{0.090} & 0.86&\conf{0.030} & 0.67&\conf{0.050}& \textbf{0.89}&\conf{0.060} & 0.64&\conf{0.059} & 0.80&\conf{0.040} & 0.60&\conf{0.055}& \textbf{0.86}&\conf{0.050}\\
   \midrule
{\textbf{Average}} & 0.75&\conf{0.050} & 0.69&\conf{0.044} & 0.66&\conf{0.056} & \textbf{0.85}&\conf{0.053} & 0.71&\conf{0.068} & 0.64&\conf{0.066} & 0.62&\conf{0.051} & \textbf{0.81}&\conf{0.045}\\
\bottomrule
\end{tabular}%
}
\vspace{-6mm}
\end{table}
\begin{table}
    \caption{\added{Execution time comparison~(in seconds -- lower is better)}} \label{tab:execution_time_nn}
\centering
    \begin{tabular}{lr@{}lr@{}lr@{}lr@{}l}
 \toprule
 Dataset  & \multicolumn{2}{c}{LIME} & \multicolumn{2}{c}{SHAP} & \multicolumn{2}{c}{MAPLE} & \multicolumn{2}{c}{BELLA}\\
  \midrule
{Auto MPG} & 2.21&\conf{0.043} & 0.04&\conf{0.001} & \textbf{0.02}&\conf{0.001} & 1.40&\conf{0.034}\\ 
{Bike} & 1.84&\conf{0.001} & 0.10&\conf{0.000} & \textbf{0.01}&\conf{0.033}& 2.70&\conf{0.000}\\
{Concrete}  & 2.10&\conf{0.034} & \textbf{0.02}&\conf{0.001} & \textbf{0.02}&\conf{0.001} & 0.92&\conf{0.023}\\
{Servo}  & 1.34&\conf{0.043} & 0.17&\conf{0.001} & \textbf{0.02}&\conf{0.001} & 2.99&\conf{0.071}\\
{Electrical}  & 3.27&\conf{0.008} & 0.13 &\conf{0.000}& \textbf{0.01}&\conf{0.000} & 2.27&\conf{0.001} \\
{Superconductors} & 21.9&\conf{0.014} & 0.17&\conf{0.020} & \textbf{0.02}&\conf{0.001} & 242.00&\conf{0.752}\\
{White Wine}  & 3.71&\conf{0.053} & 0.16&\conf{0.002} & \textbf{0.02}&\conf{0.001} & 1.66&\conf{0.014}\\
{Real Estate~~~~~} & 1.74&\conf{0.069} & \textbf{0.01}&\conf{0.000} & \textbf{0.01}&\conf{0.000} & 0.72&\conf{0.032}\\
{Wind} & 3.77&\conf{0.004} & 0.13&\conf{0.002} & \textbf{0.01}&\conf{0.000} & 1.53&\conf{0.001}\\
{CPU Activity} & 1.04&\conf{0.001} & 0.11&\conf{0.000} & \textbf{0.01}&\conf{0.000} & 1.36&\conf{0.001}\\
{Echocardiogram} & 1.83&\conf{0.003} & \textbf{0.01}&\conf{0.001} & \textbf{0.01}&\conf{0.000} & 3.35&\conf{0.002}\\
Iranian Churn & 2.83&\conf{0.037} & 0.04&\conf{0.001} & \textbf{0.02}&\conf{0.001}& 1.36&\conf{0.024}\\
  \midrule
{\textbf{Median}} & 2.16&\conf{0.024} & 0.11&\conf{0.001} &0.02&\conf{0.001} & 1.6 &\conf{0.019}\\
{\textbf{Average}} & 3.97&\conf{0.024} & 0.09&\conf{0.001} &0.02&\conf{0.001} & 21.86 &\conf{0.019}\\
\bottomrule
\end{tabular}%
\end{table}

From Tables~\ref{tab:fidelity_rmse},~\ref{tab:generality},~\ref{tab:simplicity}, and~\ref{tab:robustness}, we can see that at the same level of simplicity, BELLA provides more general, more robust, and more accurate explanations than LIME. BELLA provides less accurate explanations than SHAP and MAPLE, but at the same time, BELLA's explanations are more general, more robust, and vastly simpler. The results when the black-box model is a random forest are shown in Appendix~\ref{sec:trees}, and they do not differ much. 

\textbf{Runtime} \added{of all methods is shown in Table~\ref{tab:execution_time_nn}. On average, MAPLE and SHAP are extraordinarily fast, and LIME is slower. BELLA is, on average, 5$\times$ slower than LIME. This is due mainly to a single dataset, Superconductivity, which has a very large number of features, and on which BELLA is 11$\times$ slower than LIME. This is because, different from our competitors, BELLA is deterministic. Our method thus has to explore the whole local space. At the same time, BELLA runs in the same order of magnitude of time as LIME on average, and it remains thus competitive. On the median, BELLA is even faster than LIME.}

\textbf{The other desiderata} outlined in Section~\ref{sec:preliminaries} were determinism and verifiability (the possibility to compute the explained value from the feature values). SHAP offers none of these. Neither does LIME. While both SHAP and LIME compute linear models with feature weights, these models are not verifiable in our sense: There is no way that the user can insert the feature values of a neighboring point into these models and obtain an explained value. This is because the linear models do not operate in the original input feature space. Only MAPLE offers this verifiability. However, it relies on randomization. BELLA is thus the only approach that delivers deterministic, and verifiable explanations.

\begin{figure}
    \centering
    \includegraphics[width=0.4\linewidth]{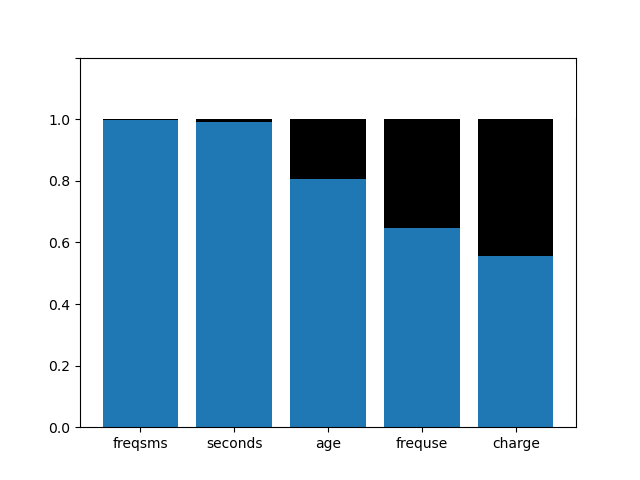}
    \caption{\added{Top 5 features of a linear model on the Iranian Churn dataset, and percentage of data points whose BELLA explanation uses each feature.}}
    \label{fig:churn_linear_model}
\end{figure}

\textbf{Verification on an interpretable model.} To confirm that the explanations provided by \method represent what the black-box model has learned, we evaluate them with regard to an already interpretable model. Instead of a black-box model, we train an Ordinary Least Square linear regression model and consider the $5$ most important features.
We then compute the explanations for each data point in the test set with our method. \method was able to recover on average $85.12\%$ of the original top-5 features across all datasets. \added{Figure~\ref{fig:churn_linear_model} shows an example of the $5$ most important features in the Iranian Churn dataset, as extracted by the linear model, and the percentage of data points for which BELLA gives an explanation with these features.} This shows that our method provides explanations that generally agree with prior beliefs, as encoded in an interpretable model. 

\section{Limitations}
\added{BELLA is a domain-independent method to produce deterministic post-hoc local explanations of tabular regression datasets. This scope entails some inherent limitations: First, as a generalist method, BELLA does not take into account domain-specific peculiarities, such as domain-specific weighting, dynamic scaling of feature distances, or domain-specific distance functions. Second, BELLA aims at simple and verifiable (and ultimately linear) explanations. With this, BELLA may fail to capture intricate, higher-order interactions of features. This problem is part of the intrinsic trade-off between simplicity and accuracy, which any post-hoc explainability approach finds itself in. Third, BELLA has to explore the whole neighborhood space without sampling, because it is deterministic. This incurs a high algorithmic cost, as our experiments show. This means that BELLA does not scale well to a large number of features, and the speed optimization of BELLA is left for future work.}

\added{There are also a number of ethical considerations: First, BELLA does not correct for biases in the data or the model. If the data contains biased samples, discriminating features, or an otherwise unprofessional selection of features or data points, these characteristics will be mirrored in BELLA's explanations. We believe, however, that this is a feature rather than a bug: BELLA’s explanations make such biases visible to the user. It would be disastrous to correct them and thus convey the impression that the model does not have them. A second concern is the problem of overtrust, where users mistakenly take BELLA's local explanation for a global correlation. That is an issue inherent to all local approaches, which would have to be countered by educating the user before using the approach. \newlyadded{Finally, BELLA measures generality by the number of training examples that were considered to compute the explanation. While BELLA minimized the average error on this set of datapoints, it is possible that some examples of this neighborhood are not well explained.}
}

\added{For all of these reasons, BELLA, as any explainability method, gains from being combined with frameworks for fairness, transparency, and accountability, as well as domain-specific adaptations.}

\section{Conclusion}
\label{conclusions}

We have presented BELLA, a deterministic approach to provide post-hoc local explanations for any regression black-box model, or indeed any static tabular dataset with a numeric variable to be explained. BELLA's 
objective function ensures accurate, general, robust, and simple explanations. Detailed experiments show that BELLA outperforms state-of-the-art approaches on these desiderata, often by a wide margin. 

Future work could investigate methods to improve the speed of BELLA. One possibility is to use other data structures to speed up the neighborhood search. Another possibility is to give up determinism and resort to sampling to further speed up the algorithm. \added{Finally, one could investigate even if BELLA could replace black box models for making a prediction in the first place, following positive experiences with linear models elsewhere~\citep{ismail2022interpretable}.} We hope that our work can open the door to research along this line and others, and ultimately make machine learning models more interpretable.

\bibliography{bibliography}
\newpage
\appendix

\section{Experiments with Random Forest as black-box model}\label{sec:trees}

In the main paper, we presented experiments using a neural network as a black-box model. Here, we show the results of experiments using a random forest of \added{100 trees} as a black-box model. The results are shown in Tables~\ref{tab:fidelity_rmse_rf},~\ref{tab:generality_rf},~\ref{tab:simplicity_rf}, and ~\ref{tab:robustness_rf}. They do not differ much from the results on the neural network black-box model.

\begin{table}[!h]
    \caption{Fidelity comparison for Random Forest as black-box model}     \label{tab:fidelity_rmse_rf}
\centering
    \begin{tabular}{lr@{}lr@{}lr@{}lr}
 \toprule
 Dataset & \multicolumn{2}{c}{LIME} & \multicolumn{2}{c}{MAPLE} & \multicolumn{2}{c}{BELLA} & SHAP  \\
  \midrule
 Auto MPG & 1.57 &\conf{0.470} & \textbf{0.95}&\conf{0.272} & 1.52&\conf{0.412} & \textbf{0.00}\\ 
 Bike & 338.50&\conf{20.92} & \textbf{61.23}&\conf{5.560} & 228.14&\conf{14.48} & \textbf{0.00}\\
 Concrete & 5.78&\conf{0.844} & \textbf{2.22}&\conf{0.341} & 5.11&\conf{0.681} & \textbf{0.00}\\
 Servo  & 0.45&\conf{0.251} & \textbf{0.18}&\conf{0.152} & 0.45&\conf{0.211} & \textbf{0.00}\\
 Electrical  & 0.01&\conf{0.002} & \textbf{0.00}&\conf{0.001} & 0.01&\conf{0.002}& \textbf{0.00} \\
 Supercond. & 30.68&\conf{1.407} & \textbf{0.96}&\conf{0.086} & 15.52&\conf{0.481} & \textbf{0.00} \\
 White Wine& 0.30&\conf{0.022} & \textbf{0.17}&\conf{0.014} & 0.28&\conf{0.023} & \textbf{0.00} \\ 
 Real Estate& 5.19&\conf{1.292} & \textbf{2.55}&\conf{0.962}& 5.04&\conf{1.503} & \textbf{0.00}\\
 Wind & 1.40&\conf{0.197} & \textbf{0.70}&\conf{0.128} & 1.17&\conf{0.184} & \textbf{0.00} \\
 CPU Activity & 12.57&\conf{1.110} & \textbf{0.63}&\conf{0.070} & 1.26&\conf{0.010} & \textbf{0.00} \\
 Echocard. & 3.31&\conf{0.510} & \textbf{1.64}&\conf{0.246} & 3.21&\conf{0.532} & \textbf{0.00} \\
 Iranian Churn  & 141.57&\conf{20.44} & \textbf{9.46}&\conf{1.783} & 17.39&\conf{2.923} & \textbf{0.00} \\
 \midrule
\textbf{Norm. avg.}& 0.07&\conf{0.010} & 0.02&\conf{0.004} & 0.05&\conf{0.008} & 0.00 \\
\bottomrule
\end{tabular}%
\end{table}

\begin{table}
\caption{Generality comparison~(\% - larger is better)}\label{tab:generality_rf}
\centering
    \begin{tabular}{lr@{}lrr@{}lr@{}l}
 \toprule
 Dataset & \multicolumn{2}{c}{LIME} & SHAP & \multicolumn{2}{c}{MAPLE} & \multicolumn{2}{c}{BELLA}\\
  \midrule
 Auto MPG & 8.35&\conf{2.342} & 0.00 & 45.41&\conf{3.423} & \textbf{72.08}&\conf{4.032}\\ 
 Bike & 1.34&\conf{0.247} & 0.00 & 6.36&\conf{0.259} & \textbf{50.96}&\conf{2.346}\\
 Concrete & 0.23&\conf{0.013} & 0.00 & 30.13&\conf{1.894} & \textbf{42.24}&\conf{4.038}\\
 Servo & 11.03&\conf{0.701} & 0.00 & 73.24 &\conf{6.034} & \textbf{84.03}&\conf{12.123}\\
 Electrical  & 0.01&\conf{0.001} & 0.00 & 7.23&\conf{0.370} & \textbf{33.24}&\conf{6.540}\\
 Supercond. & 0.01&\conf{0.002} & 0.00 & 3.47&\conf{0.720} & \textbf{67.45}&\conf{13.67}\\
 White Wine& 2.18&\conf{0.242} & 0.00 & 18.48&\conf{0.373} & \textbf{72.24}&\conf{2.439}\\ 
 Real Estate & 2.43&\conf{0.976} & 0.00 & 47.32&\conf{3.987} & \textbf{83.38}&\conf{7.320}\\
 Wind  & 0.44&\conf{0.013} & 0.00 & 12.46&\conf{0.440} & \textbf{100.0}&\conf{0.000}\\
 CPU Activity & 0.47&\conf{0.014} & 0.00 & 9.94&\conf{0.245} & \textbf{50.43}&\conf{1.430}\\
 Echocard. & 2.11&\conf{0.440} & 0.00 & 5.98&\conf{0.254} & \textbf{88.22}&\conf{5.620}\\
 Iranian Churn & 1.78&\conf{0.336} & 0.00 & 11.97&\conf{0.325} & \textbf{28.43}&\conf{1.893}\\
 \midrule
 \textbf{Average} & 2.53&\conf{0.444} & 0.00 & 22.67&\conf{1.3610} & \textbf{64.39}&\conf{3.605}\\
\bottomrule
\end{tabular}%
\end{table}

\begin{table}
    \caption{Simplicity comparison (smaller values are better)} 
 \label{tab:simplicity_rf}
\centering
    \begin{tabular}{lr@{}lr@{}lr@{}l}
 \toprule
 Dataset & \multicolumn{2}{c}{SHAP} & \multicolumn{2}{c}{MAPLE} & \multicolumn{2}{c}{BELLA/LIME}\\
  \midrule
{Auto MPG} & 9.00&\conf{0.000} & 8.85&\conf{0.210} & \textbf{3.65}&\conf{0.170}\\ 
{Bike} & 12.22&\conf{0.039} & 13.43&\conf{0.081} & \textbf{7.94}&\conf{0.103} \\
{Concrete} & 8.00&\conf{0.000} & 8.00&\conf{0.000} & \textbf{5.40}&\conf{0.301}\\
{Servo}  & 10.16&\conf{1.650} & 14.88&\conf{0.250} & \textbf{6.47}&\conf{1.290}\\
{Electrical} & 12.00&\conf{0.000} & 12.00&\conf{0.000} & \textbf{8.06}&\conf{0.201} \\
{Supercond.}  & 34.14&\conf{0.555} & 81.00&\conf{0.000} & \textbf{12.57}&\conf{0.081}\\
{White Wine} & 11.00&\conf{0.000} & 11.00&\conf{0.000} & \textbf{6.02}&\conf{0.212}\\
{Real Estate~~~~~}  & 5.00&\conf{0.000} & 5.00&\conf{0.000} & \textbf{3.95}&\conf{0.074}\\
{Wind} & 13.63&\conf{0.048} & 13.00&\conf{0.000} & \textbf{7.82}&\conf{0.155}\\
{CPU Activity} & 12.00&\conf{0.000} & 12.00&\conf{0.000} & \textbf{9.50}&\conf{0.193}\\
{Echocardiogram} & 7.36&\conf{0.108} & 8.43&\conf{0.099} & \textbf{5.23}&\conf{0.101}\\
{Iranian Churn} & 9.15&\conf{0.041} & 10.51&\conf{0.061} & \textbf{4.87}&\conf{0.161}\\
   \midrule
   {\textbf{Norm. Avg.}} & 0.85&\conf{0.009} & 0.94&\conf{0.005} & \textbf{0.53}&\conf{0.020}\\
\bottomrule
\end{tabular}
\end{table}
\clearpage
\begin{table}[!ht]
\centering
    \caption{Execution time comparison~(in seconds -- lower is better)} \label{tab:execution_time_rf}
\centering
    \begin{tabular}{lr@{}lr@{}lr@{}lr@{}l}
 \toprule
 Dataset  & \multicolumn{2}{c}{LIME} & \multicolumn{2}{c}{SHAP} & \multicolumn{2}{c}{MAPLE} & \multicolumn{2}{c}{BELLA}\\
  \midrule
{Auto MPG} & 2.32&\conf{0.040} & 0.06&\conf{0.000} & \textbf{0.01}&\conf{0.000} & 1.35&\conf{0.030}\\ 
{Bike} & 1.42&\conf{0.014} & 0.12&\conf{0.001} & \textbf{0.01}&\conf{0.000}& 3.24&\conf{0.010}\\
{Concrete}  & 2.26&\conf{0.030} & 0.04&\conf{0.000} & \textbf{0.02}&\conf{0.000} & 0.92&\conf{0.010}\\
{Servo}  & 1.42&\conf{0.061} & 0.18&\conf{0.001} & \textbf{0.02}&\conf{0.000} & 3.03&\conf{0.112}\\
{Electrical}  & 1.93&\conf{0.001} & 0.11 &\conf{0.001}& \textbf{0.01}&\conf{0.000} & 1.95&\conf{0.004} \\
{Supercond.} & 21.65&\conf{0.123} & 0.27&\conf{0.004} & \textbf{0.02}&\conf{0.001} & 243.51&\conf{0.917}\\
{White Wine}  & 3.92&\conf{0.051} & 0.20&\conf{0.001} & \textbf{0.02}&\conf{0.000} & 1.70&\conf{0.012}\\
{Real Estate~~~~~} & 1.84&\conf{0.084} & 0.02&\conf{0.000} & \textbf{0.01}&\conf{0.000} & 0.77&\conf{0.030}\\
{Wind} & 2.24&\conf{0.001} & 0.13&\conf{0.001} & \textbf{0.01}&\conf{0.000} & 1.45&\conf{0.003}\\
{CPU Activity} & 1.87&\conf{0.001} & 0.12&\conf{0.001} & \textbf{0.01}&\conf{0.000} & 1.66&\conf{0.021}\\
{Echocardiogram} & 1.17&\conf{0.001} & 0.02&\conf{0.001} & \textbf{0.01}&\conf{0.001} & 2.57&\conf{0.532}\\
Iranian Churn & 3.00&\conf{0.041} & 0.08&\conf{0.001} & \textbf{0.02}&\conf{0.000}& 1.40&\conf{0.012}\\
\bottomrule
\end{tabular}%
\end{table}

\begin{table}[!h]
    \caption{Robustness comparison~(0 to 1 -- larger is better)} \label{tab:robustness_rf}
\centering
\scalebox{0.8}{%
    \begin{tabular}{lr@{}lr@{}lr@{}lr@{}lr@{}lr@{}lr@{}lr@{}lr@{}lr@{}lr@{}lr@{}l}
 \toprule
 Dataset  & \multicolumn{8}{c}{\textbf{Number of Neighbours = 5}}& \multicolumn{8}{c}{\textbf{Number of Neighbours = 20}} \\
 \cmidrule(lr){2-9} \cmidrule(lr){10-17}
          & \multicolumn{2}{c}{LIME} & \multicolumn{2}{c}{SHAP} & \multicolumn{2}{c}{MAPLE} & \multicolumn{2}{c}{BELLA} & \multicolumn{2}{c}{LIME} & \multicolumn{2}{c}{SHAP} & \multicolumn{2}{c}{MAPLE} & \multicolumn{2}{c}{BELLA} \\
  \midrule
{Auto MPG} & 0.90&\conf{0.040} & 0.68&\conf{0.083} & 0.70&\conf{0.050} & \textbf{0.96}&\conf{0.020} & 0.83&\conf{0.024} & 0.58&\conf{0.073} & 0.65&\conf{0.047} & \textbf{0.93}&\conf{0.028}\\ 
{Bike} & 0.69&\conf{0.040} & 0.63&\conf{0.030} & 0.47&\conf{0.060}& \textbf{0.83}&\conf{0.060} & 0.67&\conf{0.047} & 0.62&\conf{0.037} & 0.48&\conf{0.050}& \textbf{0.85}&\conf{0.048}\\
{Concrete} & \textbf{0.81}&\conf{0.050} & 0.73&\conf{0.050} & 0.76&\conf{0.070} & 0.79&\conf{0.040}& 0.67&\conf{0.067} & 0.58&\conf{0.046} & 0.65&\conf{0.039} & \textbf{0.73}&\conf{0.070}\\
{Servo} & \textbf{0.85}&\conf{0.050} & 0.46&\conf{0.040} & 0.63&\conf{0.110} & 0.74&\conf{0.040} & \textbf{0.80}&\conf{0.033} & 0.59&\conf{0.041} & 0.62&\conf{0.1274} & 0.76&\conf{0.1347}\\
{Electrical} & 0.82&\conf{0.034} & 0.55 &\conf{0.037}& 0.62&\conf{0.046} & \textbf{0.88}&\conf{0.028} & 0.76&\conf{0.036} & 0.50 &\conf{0.022}&0.58&\conf{0.028} & \textbf{0.88}&\conf{0.026}\\
{Supercond.} & 0.88&\conf{0.013} & 0.69&\conf{0.044} & 0.68&\conf{0.091} & \textbf{0.94}&\conf{0.035} & 0.86&\conf{0.072} & 0.73&\conf{0.013} & 0.68&\conf{0.018} & \textbf{0.95}&\conf{0.067}\\
{White Wine} & 0.74&\conf{0.070} & 0.58&\conf{0.080} & 0.62&\conf{0.070} & \textbf{0.87}&\conf{0.060} & 0.66&\conf{0.061} & 0.46&\conf{0.033} & 0.53&\conf{0.045} & \textbf{0.83}&\conf{0.038}\\
{Real Estate~~~~~} & 0.73&\conf{0.080} & 0.71&\conf{0.071} & 0.76&\conf{0.059} & \textbf{0.94}&\conf{0.060} & 0.66&\conf{0.052} & 0.64&\conf{0.068} & 0.67&\conf{0.038} & \textbf{0.82}&\conf{0.090}\\
{Wind} & 0.64&\conf{0.043} & 0.61&\conf{0.043} & 0.66&\conf{0.022} & \textbf{0.99}&\conf{0.009} & 0.68&\conf{0.050} & 0.61&\conf{0.041} & 0.61&\conf{0.025} & \textbf{0.99}&\conf{0.008}\\
{CPU Activity} & 0.58&\conf{0.080} & 0.73&\conf{0.060} & 0.67&\conf{0.070} & \textbf{0.82}&\conf{0.060} & 0.55&\conf{0.095} & 0.66&\conf{0.052} & 0.62&\conf{0.051} & \textbf{0.79}&\conf{0.044}\\
{Echocardiogram} & 0.83&\conf{0.039} & 0.59&\conf{0.040} & 0.52&\conf{0.046} & \textbf{0.97}&\conf{0.016} & 0.80&\conf{0.036} & 0.57&\conf{0.026} & 0.45&\conf{0.031} & \textbf{0.96}&\conf{0.032}\\
Iranian Churn & 0.77&\conf{0.06} & 0.83&\conf{0.020} & 0.70&\conf{0.050}& \textbf{0.86}&\conf{0.049} & 0.76&\conf{0.058} & 0.77&\conf{0.054} & 0.62&\conf{0.066}& \textbf{0.84}&\conf{0.048}\\
   \midrule
{\textbf{Average}} & 0.77&\conf{0.050} & 0.65&\conf{0.049} & 0.65&\conf{0.062} & \textbf{0.88}&\conf{0.040} & 0.73&\conf{0.053} & 0.61&\conf{0.042} & 0.60&\conf{0.047} & \textbf{0.86}&\conf{0.053}\\
\bottomrule
\end{tabular}%
}
\end{table}

\end{document}

%% file: bad_explanation.tikz


\begin{tikzpicture}
  \draw[->, line width=2.125984250629301, draw=black] (0.933333, 0.966667) -- (0.933333, 6.229167);
  \draw[->, line width=2.125984250629301, draw=black] (0.933333, 0.966667) -- (8.462500, 0.966667);
    \node[right,font=, line width=0.0, color=black, font=\fontsize{21.259843}{21.259843}\selectfont] at (0.083333, 5.554167) {$y$\vphantom{gI}};
;
    \node[right,font=, line width=0.0, color=black, font=\fontsize{21.259843}{21.259843}\selectfont] at (7.541667, 0.495833) {$x$\vphantom{gI}};
;
  \path[line width=0.7086614168764338, fill=black] (1.537500, 1.762500) ellipse (0.083333 and 0.083333);
  \path[line width=0.7086614168764338, fill=black] (1.929167, 1.945833) ellipse (0.083333 and 0.083333);
  \path[line width=0.7086614168764338, fill=black] (1.754167, 2.387500) ellipse (0.083333 and 0.083333);
  \path[line width=0.7086614168764338, fill=black] (2.037500, 2.470833) ellipse (0.083333 and 0.083333);
  \path[line width=0.7086614168764338, fill=black] (2.120833, 2.937500) ellipse (0.083333 and 0.083333);
  \path[line width=0.7086614168764338, fill=black] (2.504167, 3.179167) ellipse (0.083333 and 0.083333);
  \path[line width=0.7086614168764338, fill=black] (2.654167, 3.529167) ellipse (0.083333 and 0.083333);
  \path[line width=0.7086614168764338, fill=black] (3.012500, 3.579167) ellipse (0.083333 and 0.083333);
  \path[line width=0.7086614168764338, fill=black] (2.712500, 3.979167) ellipse (0.083333 and 0.083333);
  \path[line width=0.7086614168764338, fill=black] (3.162500, 4.137500) ellipse (0.083333 and 0.083333);
  \path[line width=0.7086614168764338, fill=black] (3.412500, 4.554167) ellipse (0.083333 and 0.083333);
  \path[line width=0.7086614168764338, fill=black] (3.637500, 4.420833) ellipse (0.083333 and 0.083333);
  \path[line width=0.7086614168764338, fill=black] (3.845833, 3.937500) ellipse (0.083333 and 0.083333);
  \path[line width=0.7086614168764338, fill=black] (3.987500, 4.254167) ellipse (0.083333 and 0.083333);
  \path[line width=0.7086614168764338, fill=black] (4.254167, 3.904167) ellipse (0.083333 and 0.083333);
  \path[line width=0.7086614168764338, fill=black] (4.045833, 3.695833) ellipse (0.083333 and 0.083333);
  \path[line width=0.7086614168764338, fill=black] (4.479167, 3.412500) ellipse (0.083333 and 0.083333);
  \path[line width=0.7086614168764338, fill=black] (4.662500, 3.629167) ellipse (0.083333 and 0.083333);
  \path[line width=0.7086614168764338, fill=black] (4.554167, 3.054167) ellipse (0.083333 and 0.083333);
  \path[line width=0.7086614168764338, fill=black] (4.820833, 3.120833) ellipse (0.083333 and 0.083333);
  \path[line width=0.7086614168764338, fill=black] (5.020833, 2.645833) ellipse (0.083333 and 0.083333);
  \path[line width=0.7086614168764338, fill=black] (5.254167, 2.279167) ellipse (0.083333 and 0.083333);
  \path[line width=0.7086614168764338, fill=black] (5.387500, 2.604167) ellipse (0.083333 and 0.083333);
  \path[line width=0.7086614168764338, fill=black] (5.620833, 2.962500) ellipse (0.083333 and 0.083333);
  \path[line width=0.7086614168764338, fill=black] (5.762500, 3.295833) ellipse (0.083333 and 0.083333);
  \path[line width=0.7086614168764338, fill=black] (6.204167, 3.087500) ellipse (0.083333 and 0.083333);
  \definecolor{athblue}{RGB}{84, 120, 183}
  \definecolor{athgreen}{RGB}{51, 187, 51}
  \path[line width=0.7086614168764338, fill=black] (5.895833, 3.687500) ellipse (0.083333 and 0.083333);
  \path[line width=0.7086614168764338, fill=black] (6.245833, 3.887500) ellipse (0.083333 and 0.083333);
  \path[line width=0.7086614168764338, fill=black] (5.995833, 4.037500) ellipse (0.083333 and 0.083333);
  \path[line width=0.7086614168764338, fill=black] (6.370833, 4.154167) ellipse (0.083333 and 0.083333);
  \path[line width=0.7086614168764338, fill=black] (6.479167, 4.495833) ellipse (0.083333 and 0.083333);
  \path[line width=0.7086614168764338, fill=black] (6.695833, 4.262500) ellipse (0.083333 and 0.083333);
  \path[line width=0.7086614168764338, fill=black] (6.770833, 4.645833) ellipse (0.083333 and 0.083333);
  \path[line width=0.7086614168764338, fill=black] (7.079167, 4.620833) ellipse (0.083333 and 0.083333);
  \path[line width=0.7086614168764338, fill=black] (7.329167, 4.295833) ellipse (0.083333 and 0.083333);
  \path[line width=0.7086614168764338, fill=black] (7.929167, 3.929167) ellipse (0.083333 and 0.083333);
  \path[line width=0.7086614168764338, fill=black] (7.579167, 4.020833) ellipse (0.083333 and 0.083333);
  \path[line width=0.7086614168764338, fill=black] (7.737500, 3.629167) ellipse (0.083333 and 0.083333);
  \path[line width=0.7086614168764338, fill=black] (8.029167, 3.395833) ellipse (0.083333 and 0.083333);
  \path[line width=0.7086614168764338, fill=black] (1.729167, 1.479167) ellipse (0.083333 and 0.083333);
  \path[line width=0.7086614168764338, fill=black] (1.270833, 1.304167) ellipse (0.083333 and 0.083333);
  \draw[line width=2.125984250629301, draw=athblue] (5.916667, 4.358333) -- (6.266667, 2.616667);
  \path[line width=0.7086614168764338, fill=athgreen ] (6.070833, 3.454167) ellipse (0.166667 and 0.166667);
\end{tikzpicture}

%% file: good_explanation.tikz


\begin{tikzpicture}
  \draw[->, line width=2.125984250629301, draw=black] (0.933333, 0.966667) -- (0.933333, 6.229167);
  \draw[->, line width=2.125984250629301, draw=black] (0.933333, 0.966667) -- (8.462500, 0.966667);
    \node[right,font=, line width=0.0, color=black, font=\fontsize{21.259843}{21.259843}\selectfont] at (0.083333, 5.554167) {$y$\vphantom{gI}};
;
    \node[right,font=, line width=0.0, color=black, font=\fontsize{21.259843}{21.259843}\selectfont] at (7.541667, 0.495833) {$x$\vphantom{gI}};
;
  \path[line width=0.7086614168764338, fill=black] (1.537500, 1.762500) ellipse (0.083333 and 0.083333);
  \path[line width=0.7086614168764338, fill=black] (1.929167, 1.945833) ellipse (0.083333 and 0.083333);
  \path[line width=0.7086614168764338, fill=black] (1.754167, 2.387500) ellipse (0.083333 and 0.083333);
  \path[line width=0.7086614168764338, fill=black] (2.037500, 2.470833) ellipse (0.083333 and 0.083333);
  \path[line width=0.7086614168764338, fill=black] (2.120833, 2.937500) ellipse (0.083333 and 0.083333);
  \path[line width=0.7086614168764338, fill=black] (2.504167, 3.179167) ellipse (0.083333 and 0.083333);
  \path[line width=0.7086614168764338, fill=black] (2.654167, 3.529167) ellipse (0.083333 and 0.083333);
  \path[line width=0.7086614168764338, fill=black] (3.012500, 3.579167) ellipse (0.083333 and 0.083333);
  \path[line width=0.7086614168764338, fill=black] (2.712500, 3.979167) ellipse (0.083333 and 0.083333);
  \path[line width=0.7086614168764338, fill=black] (3.162500, 4.137500) ellipse (0.083333 and 0.083333);
  \path[line width=0.7086614168764338, fill=black] (3.412500, 4.554167) ellipse (0.083333 and 0.083333);
  \path[line width=0.7086614168764338, fill=black] (3.637500, 4.420833) ellipse (0.083333 and 0.083333);
  \path[line width=0.7086614168764338, fill=black] (3.845833, 3.937500) ellipse (0.083333 and 0.083333);
  \path[line width=0.7086614168764338, fill=black] (3.987500, 4.254167) ellipse (0.083333 and 0.083333);
  \path[line width=0.7086614168764338, fill=black] (4.254167, 3.904167) ellipse (0.083333 and 0.083333);
  \path[line width=0.7086614168764338, fill=black] (4.045833, 3.695833) ellipse (0.083333 and 0.083333);
  \path[line width=0.7086614168764338, fill=black] (4.479167, 3.412500) ellipse (0.083333 and 0.083333);
  \path[line width=0.7086614168764338, fill=black] (4.662500, 3.629167) ellipse (0.083333 and 0.083333);
  \path[line width=0.7086614168764338, fill=black] (4.554167, 3.054167) ellipse (0.083333 and 0.083333);
  \path[line width=0.7086614168764338, fill=black] (4.820833, 3.120833) ellipse (0.083333 and 0.083333);
  \path[line width=0.7086614168764338, fill=black] (5.020833, 2.645833) ellipse (0.083333 and 0.083333);
  \path[line width=0.7086614168764338, fill=black] (5.254167, 2.279167) ellipse (0.083333 and 0.083333);
  \path[line width=0.7086614168764338, fill=black] (5.387500, 2.604167) ellipse (0.083333 and 0.083333);
  \path[line width=0.7086614168764338, fill=black] (5.620833, 2.962500) ellipse (0.083333 and 0.083333);
  \path[line width=0.7086614168764338, fill=black] (5.762500, 3.295833) ellipse (0.083333 and 0.083333);
  \path[line width=0.7086614168764338, fill=black] (6.204167, 3.087500) ellipse (0.083333 and 0.083333);
    \definecolor{athblue}{RGB}{84, 120, 183}
  \definecolor{athgreen}{RGB}{51, 187, 51}
  \path[line width=0.7086614168764338, fill=black] (5.895833, 3.687500) ellipse (0.083333 and 0.083333);
  \path[line width=0.7086614168764338, fill=black] (6.245833, 3.887500) ellipse (0.083333 and 0.083333);
  \path[line width=0.7086614168764338, fill=black] (5.995833, 4.037500) ellipse (0.083333 and 0.083333);
  \path[line width=0.7086614168764338, fill=black] (6.370833, 4.154167) ellipse (0.083333 and 0.083333);
  \path[line width=0.7086614168764338, fill=black] (6.479167, 4.495833) ellipse (0.083333 and 0.083333);
  \path[line width=0.7086614168764338, fill=black] (6.695833, 4.262500) ellipse (0.083333 and 0.083333);
  \path[line width=0.7086614168764338, fill=black] (6.770833, 4.645833) ellipse (0.083333 and 0.083333);
  \path[line width=0.7086614168764338, fill=black] (7.079167, 4.620833) ellipse (0.083333 and 0.083333);
  \path[line width=0.7086614168764338, fill=black] (7.329167, 4.295833) ellipse (0.083333 and 0.083333);
  \path[line width=0.7086614168764338, fill=black] (7.929167, 3.929167) ellipse (0.083333 and 0.083333);
  \path[line width=0.7086614168764338, fill=black] (7.579167, 4.020833) ellipse (0.083333 and 0.083333);
  \path[line width=0.7086614168764338, fill=black] (7.737500, 3.629167) ellipse (0.083333 and 0.083333);
  \path[line width=0.7086614168764338, fill=black] (8.029167, 3.395833) ellipse (0.083333 and 0.083333);
  \path[line width=0.7086614168764338, fill=black] (1.729167, 1.479167) ellipse (0.083333 and 0.083333);
  \path[line width=0.7086614168764338, fill=black] (1.270833, 1.304167) ellipse (0.083333 and 0.083333);
  \draw[line width=2.125984250629301, draw=athblue] (7.175000, 5.316667) -- (5.091667, 1.858333);
  \path[line width=0.7086614168764338, fill=athgreen ] (6.070833, 3.454167) ellipse (0.166667 and 0.166667);
\end{tikzpicture}

%% file: exp_example2.tikz



\begin{minipage}{1\columnwidth}
\centering
\includegraphics[width=0.6\textwidth]{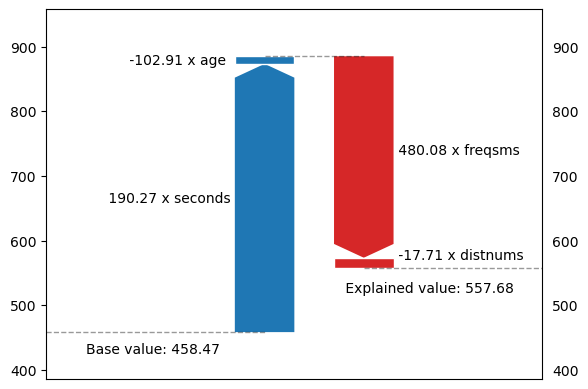}\ \\

The value predicted by the model is $\textbf{551}$ and the explained value is $\textbf{557}$. This explanation applies to $\textbf{476}$ other instances.
\end{minipage}